\documentclass[runningheads]{llncs}
\usepackage[T1]{fontenc}
\usepackage{graphicx,verbatim}
\usepackage{multirow}

\begin{document}
\title{ReSW-VL: Representation Learning for Surgical Workflow Analysis Using Vision-Language Model}

\titlerunning{ReSW-VL: Representation Learning for Surgical Workflow Analysis}
%
\author{Satoshi Kondo\inst{1}\orcidID{0000-0002-4941-4920}}
\authorrunning{S. Kondo}
%
\institute{Muroran Institute of Technology, Hokkaido, Japan\\
}

\maketitle              
\begin{abstract}

Surgical phase recognition from video is a technology that automatically classifies the progress of a surgical procedure and has a wide range of potential applications, including real-time surgical support, optimization of medical resources, training and skill assessment, and safety improvement. Recent advances in surgical phase recognition technology have focused primarily on Transform-based methods, although methods that extract spatial features from individual frames using a CNN and video features from the resulting time series of spatial features using time series modeling have shown high performance. However, there remains a paucity of research on training methods for CNNs employed for feature extraction or representation learning in surgical phase recognition. In this study, we propose a method for representation learning in surgical workflow analysis using a vision-language model (ReSW-VL). Our proposed method involves fine-tuning the image encoder of a CLIP (Convolutional Language Image Model) vision-language model using prompt learning for surgical phase recognition. The experimental results on three surgical phase recognition datasets demonstrate the effectiveness of the proposed method in comparison to conventional methods.


\end{abstract}

\section{Introduction}

The implementation of surgical phase recognition technology entails the analysis of surgical videos, thereby facilitating the automatic classification of surgical progress. This technology confers numerous advantages, encompassing real-time surgical assistance, optimized resource utilization, educational prospects, skill evaluation, and enhanced safety measures. For instance, real-time progress monitoring during surgical procedures enables medical personnel to expeditiously and precisely ascertain the subsequent step, thereby enhancing surgical efficiency and safety. Furthermore, the analysis of surgical records facilitates the effective allocation of medical resources, as well as the training and assessment of medical students and residents, which is expected to improve the quality of surgery and reduce the burden on medical personnel.

Recent advances in surgical phase recognition using deep learning have been reported~\cite{demir2023deep,li2024deep}. Surgical phase recognition is a process performed on videos; therefore, the key point is how to capture the spatio-temporal features necessary for phase recognition from image sequences. Surgical phase recognition technology generally extracts spatial features for each frame in the video, and then extracts the features of the video by time-series modeling from the time-series of the obtained spatial features. Convolutional Neural Networks (CNNs), exemplified by ResNet~\cite{he2016deep} and Vision Transformer (ViT)~\cite{dosovitskiy2020image}, are frequently employed for the extraction of spatial features. Initially, Recurrent Neural Networks (RNNs)~\cite{mikolov2010recurrent} and Long Short-Term Memory (LSTM) networks~\cite{hochreiter1997long} were utilized for time series modeling, followed by Temporal Convolutional Networks (TCNs)~\cite{lea2016temporal} and Transformers~\cite{vaswani2017attention}. In recent years, studies have emerged that propose the application of Transformers for both spatial feature extraction and time series modeling. However, it is still reported that the combination of CNNs and time series modeling yields superior performance~\cite{rivoir2024pitfalls}.

The combination of CNN and time series modeling is generally trained in two stages. In the first stage, the CNN is trained as a spatial feature extractor. In the second stage, the time series modeling is trained. In the second stage, the network parameters of the CNN remain fixed. The primary function of the CNNs in the first stage is to classify phases, and while the enhancement of phase recognition performance through the acquisition of effective spatial features with CNNs is paramount, there is a paucity of studies that have thoroughly examined the training methodologies of CNNs in phase recognition.

In this paper, we propose a novel representation learning method for extracting spatial features from images, with the objective of facilitating surgical phase recognition. The proposed method draws inspiration from extant literature~\cite{li2022ordinalclip}. We utilize CLIP~\cite{radford2021learning}, a visual-language model, to perform prompt learning and image encoder fine tuning. The resulting image encoder functions as a feature extractor, and in conjunction with time series modeling, enables surgical phase recognition.

The contributions of this study are as follows:
\begin{itemize}
\item This is one of the few studies on representation learning for surgical phase recognition, and it is the first method to use the vision-language model and prompt learning.
\item We show that the proposed method outperforms conventional methods on three widely used surgical phase recognition datasets.
\end{itemize}
The code will be made publicly available if the paper is accepted.

\section{Related Works}

Table~\ref{tab1} shows representative prior arts on deep learning-based surgical phase recognition methods. As illustrated in Table~\ref{tab1}, conventional surgical phase recognition methodologies predominantly entail the extraction of spatial features from individual frames within a video sequence. These spatial features are then utilized to derive video features through the application of time series modeling techniques. As illustrated in Table~\ref{tab1}, CNN (ResNet-50) and Vision Transformer (ViT-B/16) are frequently employed for spatial feature extraction, while LSTM, TCN, and Transformer are often utilized for time series modeling.

{
\tabcolsep = 3pt
\begin{table}[t]
\caption{Representative prior arts on deep learning-based surgical phase recognition methods.}
\label{tab1}
\centering
\begin{tabular}{l|c|c|c}
\hline
\multicolumn{1}{c|}{Reference} & Year & Spatial feature extractor & Temporal modeling \\
\hline
Jin et al.~\cite{jin2017sv} (SV-RCNet) & 2018 & ResNet-50 & LSTM \\
Czempiel et al.~\cite{czempiel2020tecno} (TeCNO) & 2020 & ResNet-50 & TCN \\
Gao et al.~\cite{gao2021trans} (Trans-SVNet) & 2021 & ResNet-50 & TCN, Transformer \\
Czempiel et al.~\cite{czempiel2021opera} (Opera) & 2021 & ResNet-50 & Transformer \\
Zou et al.~\cite{zou2023arst} (ARST) & 2022 & ResNet-50 & TCN, Transformer \\
Liu et al.~\cite{liu2023skit} (SKiT) & 2023 & ViT-B/16 & Transformer \\
Liu et al.~\cite{liu2025lovit} (LoViT) & 2025 & ViT-B/16 & Transformer \\\hline
\end{tabular}
\end{table}
}

In the context of training feature extractors, CNNs, and Vision Transformers for surgical phase recognition, a two-stage approach has been employed, wherein the feature extractors are fine-tuned with parameters initially trained on the ImageNet dataset~\cite{deng2009imagenet}. This two-stage training method has been shown to enhance the predictive capabilities of surgical phases.

The training of the feature extractor in the first stage can be regarded as representation learning from images. Consequently, the acquisition of suitable spatial features for surgical phase recognition is crucial for enhancing the performance of surgical phase recognition. However, there is a paucity of studies that have thoroughly examined the training methodologies of feature extractors in the context of surgical phase recognition. In the subsequent section, we propose a methodology to address this knowledge gap.

\section{Propose Method}

In this paper, we propose a novel representation learning methodology for the extraction of spatial features from images, with the objective of facilitating surgical phase recognition. The proposed methodology employs CLIP, a visual-language model, to facilitate prompt learning and image encoder fine-tuning. We have designated our proposed methodology as ReSW-VL, an acronym for ``\underline{Re}presentation learning in \underline{S}urgical \underline{W}orkflow analysis using a \underline{V}ision-\underline{L}anguage model.''

First, we describe the representation of surgical phases. In some surgical phase recognition datasets, the surgical phases are represented as strings. For example, in the Cholec80 dataset~\cite{twinanda2016endonet}, the seven phases are defined as ``preparation'', ``calot triangle dissection'', ``clipping and cutting'', ``gallbladder dissection'', ``gallbladder packing'', ``cleaning and coagulation'', and ``gallbladder retraction''. These are converted to numbers defined by the dataset. If the number of surgical phases is $P$, then the number of surgical phases $p$ takes the value $1 \le p \le P$.

\begin{figure}[t]
\centering
\includegraphics[scale=0.75]{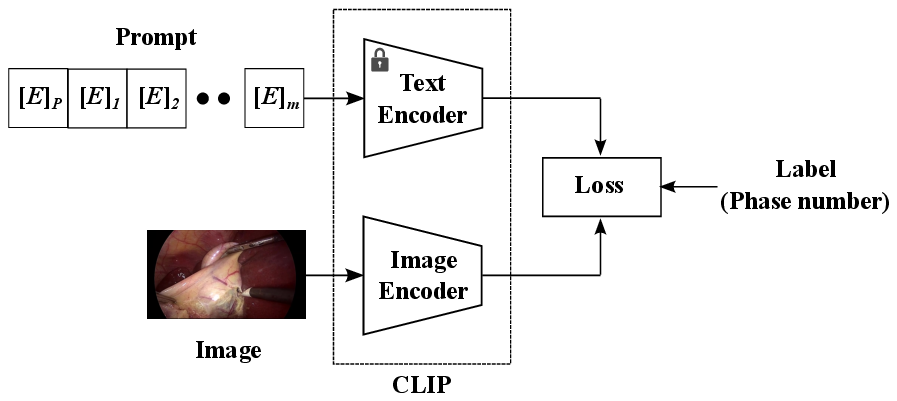}
\caption{Overview of the first stage of the proposed method.} \label{fig1}
\end{figure}

As illustrated in Figure~\ref{fig1}, the proposed method involves the utilization of both the image encoder and the text encoder, both of which are derived from CLIP. It is noteworthy that the text encoder does not undergo fine-tuning, while the image encoder undergoes fine-tuning and the prompts are learned. The first token of the prompt ([$E$]$_{P}$) corresponds to be the surgery phase number, with the subsequent $m$ tokens being obtained through learning. It is anticipated that prompt learning will yield prompts such as ``$p$-th surgical phase image in a video.'' 

The image encoder converts the input image into a $d$-dimensional feature vector. The text encoder produces a $P \times d$-dimensional feature vector corresponding to $P$ prompts. Finally, by computing the inner product of the image feature vector and the $P$ text feature vectors, a $P$-dimensional logit is obtained. The cross-entropy loss between the logit and the ground truth labels for the surgery phase is computed, and training continues by updating the image encoder and prompts.

The proposed method employs two distinct approaches to ascertain the first token of the prompts ([$E$]$_{P}$). The first approach is to determine the first token independently, without any relation to the $P$ first tokens, which is referred to as ReSW-VLi (independent). The second approach involves relating the $P$ first tokens to each other. For instance, in the Cholec80 dataset, the phase numbers are considered sequential due to the fact that, in many cases, the surgical phase numbers are incremented over time. Preserving the order in the embedding space formed by the text encoder is therefore essential. To that end, we adopt the method proposed in \cite{li2022ordinalclip}. In this approach, $n (\le P)$ first tokens are set as references, and the interpolation of these tokens is used to represent the remaining first tokens. To illustrate this approach, consider a scenario where $n = 3$. In this case, the first tokens with phase numbers $1$, $P/2$, and $P$ are learned, and the interpolation of the first tokens for the remaining phase numbers is derived from these three learned tokens. This method is referred to as ReSW-VLo (order).

\begin{figure}[t]
\centering
\includegraphics[scale=0.75]{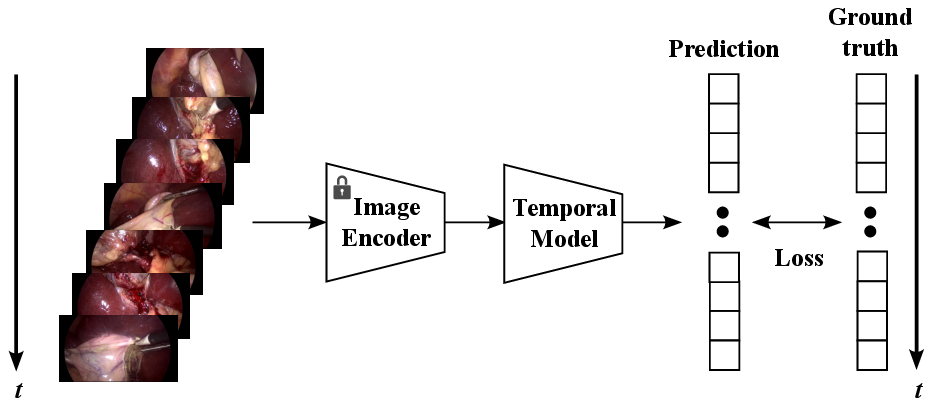}
\caption{Overview of the second stage of the proposed method.} \label{fig2}
\end{figure}

Following the refinement of the image encoder and the completion of prompt training, the subsequent stage of training the temporal model is initiated. As illustrated in Figure~\ref{fig2}, this stage involves the freezing of the image encoder, with the temporal model undergoing training through the utilization of video inputs. The temporal model can be constructed employing LSTM, TCN, or Transformer architectures.

\section{Experiments and Discussion}

Three widely utilized datasets, Cholec80~\cite{twinanda2016endonet}, Autolaparo~\cite{wang2022autolaparo}, and m2cai16~\cite{stauder2016tum,twinanda2016endonet}, were employed in the experiments. These datasets consist of laparoscopic surgical videos. The number of videos and other pertinent information is enumerated in Table~\ref{tab2}.

{
\tabcolsep = 3pt
\begin{table}[b]
\caption{Number of videos, phases, training data, validation data and test data in each dataset utilized in the experiments.}
\label{tab2}
\centering
\begin{tabular}{c|c|c|c|c|c}
\hline
Dataset & Videos & Phases & Training data & Validation data & Test data \\\hline
Cholec80 & 80 & 7 & 32 & 8 & 40 \\
Autolaparo & 21 & 7 & 10 & 4 & 7 \\ 
m2cai16 & 41 & 8 & 20 & 7 & 14 \\\hline
\end{tabular}
\end{table}
}

ResNet-50 was utilized as the image encoder for CLIP during the first stage of the process. All video in the datasets were downsampled to 1 fps. For the training phase, each frame was resized to $256 \times 256$ pixels, then augmented (horizontal flip, translation, scaling, rotation), and cropped to $224 \times 224$ pixels. During the evaluation and testing phases, each frame was resized to $224 \times 224$ pixels. The loss function was a weighted cross-entropy loss, in which the weights were calculated by median frequency balancing~\cite{eigen2015predicting}. The optimization method employed was AdamW~\cite{loshchilov2017decoupled}, the scheduler for the learning rate was cosine annealing, and the number of epochs was set to $50$. The optimal learning rate was determined by utilizing the validation data within the range of $5 \times 10^{-6}$ to $5 \times 10^{-4}$. Based on preliminary experiments, $n$ in ReSW-VLo was set to 3. The first stage of processing involves the conversion of each frame in the video into a 1024-dimensional feature vector, that is, $d=1024$.

In the second stage, the time series modeling employed a causal TCN~\cite{czempiel2020tecno}, where the TCN loss function corresponded to a weighted cross-entropy loss calculated by median frequency balancing~\cite{eigen2015predicting}. The configuration of the TCN stage was set to $1$, the number of layers to $8$, and the feature vector to $256$ dimensions. The number of epochs was set to 25. The optimizer, the scheduler, and the method to select the optimal learning rate were identical to the setting for the first stage.

In order to facilitate a meaningful comparison, ResNet-50 was selected as the spatial feature extractor in the first stage. Prior to this study, ResNet-50 was pre-trained with the ImageNet dataset and subsequently fine-tuned with the experimental dataset. The hyperparameters utilized in training are identical to those employed for the proposed method. The training in the second stage is identical to the proposed methods. This method is henceforth referred to as the ``conventional method.''

The following evaluation metrics were employed: accuracy, precision, recall, Jaccard index, and F1 score. Precision, recall, and Jaccard index were calculated as phase-wise video-wise evaluation metrics~\cite{funke2023metrics}. F1 score was calculated as the harmonic mean of the mean precision and the mean recall. Relaxed evaluation metrics were not utilized.

As demonstrated in Table~\ref{tab3}, the evaluation results for the test data substantiate the efficacy of the proposed methods. The analysis presented in Table~\ref{tab3} indicates that the proposed method consistently outperforms the conventional approach across all datasets. The enhancement achieved by the proposed methods ranges from approximately 1.0 to 4.3 percentage points for accuracy, from approximately 3.7 to 6.1 percentage points for the Jaccard index, and from approximately 2.4 to 3.9 percentage points for the F1 score.

The proposed methods, ReSW-VLi and ReSW-VLo, showed different performance depending on the dataset. Comparing the F1 scores, ReSW-VLo performed better on the Cholec80 dataset, while ReSW-VLi performed better on the Autolaparo dataset. For the M2cai dataset, the performance was almost identical.  In the Cholec80 dataset, the surgical phase number increases by one over time in many cases. Therefore, the assumption of sequentiality of phase numbers in ReSW-VLo is valid and the performance is considered higher than that of ReSW-VLi. In the Autolaparo dataset, the surgical phase number sometimes increases with time and then decreases again, which does not satisfy the sequentiality condition. Therefore, ReSW-VLi is considered to have higher performance than ReSW-VLo. Therefore, it is necessary to decide whether to use ReSW-VLi or ReSW-VLo depending on the characteristics of the dataset.

{
\tabcolsep = 3pt
\begin{table}[t]
\caption{A quantitative comparison of the conventional method and the proposed methods is presented. The average values as phase-wise video-wise evaluation metrics (in percentages) and standard deviations over phases ($\pm$) are shown. The bold font indicates the highest average value of the three methods.}
\label{tab3}
\centering
\begin{tabular}{c|c|c|c|c|c|c}
\hline
& \begin{tabular}{c} Spatial \\ feature \\extractor \end{tabular}
& Accuracy & Precision & Recall & Jaccard & F1 \\\hline
\multirow{3}{*}{\rotatebox{90}{Cholec80\!}} & Conventional 
& 82.58$\pm$9.09 & 76.90$\pm$9.30 & 78.92$\pm$8.47 & 62.36$\pm$12.06 & 77.89\rule[0mm]{0mm}{5mm} \\
& ReSW-VLi
& 84.20$\pm$8.70 & 79.48$\pm$9.57 & 77.94$\pm$10.44 & 63.47$\pm$10.73 & 78.70 \\
& ReSW-VLo
& {\bf 85.61}$\pm$9.24 & {\bf 81.11}$\pm$8.13 & {\bf 82.52}$\pm$5.44 & {\bf 68.45}$\pm$12.26 & {\bf 81.81}\rule[-3mm]{0mm}{4mm} \\\hline
\multirow{3}{*}{\rotatebox{90}{\scriptsize  Autolaparo\!\!}} & Conventional 
& 75.68$\pm$12.90 & 73.03$\pm$9.58 & 71.01$\pm$12.66 & 57.37$\pm$14.54 & 72.01\rule[0mm]{0mm}{5mm}\\
& ReSW-VLi 
& 75.69$\pm$16.25 & {\bf 75.82}$\pm$12.00 & {\bf 73.06}$\pm$11.77 & {\bf 61.02}$\pm$15.36 & {\bf 74.69} \\
& ReSW-VLo
& {\bf 76.65}$\pm$12.67 & 73.08$\pm$12.24 & 72.45$\pm$11.27 & 59.40$\pm$13.73 & 72.76\rule[-3mm]{0mm}{4mm}\\\hline
\multirow{3}{*}{\rotatebox{90}{m2cai16}} 
& Conventional 
& 70.93$\pm$13.86 & 71.55$\pm$7.50 & 73.70$\pm$10.64 & 53.43$\pm$11.97 & 72.61\rule[0mm]{0mm}{5mm} \\
& ReSW-VLi  
& {\bf 75.25}$\pm$9.91 & 73.60$\pm$7.27 & {\bf 76.41}$\pm$9.70 & {\bf 59.15}$\pm$10.38 & {\bf 74.98} \\
& ReSW-VLo 
& 73.02$\pm$12.31 & {\bf 76.71}$\pm$8.01 & 73.15$\pm$9.76 & 56.93$\pm$11.47 & 74.89\rule[-3mm]{0mm}{4mm} \\\hline
\end{tabular}
\end{table}
}

\begin{figure}[t]
\centering
\includegraphics[scale=0.65]{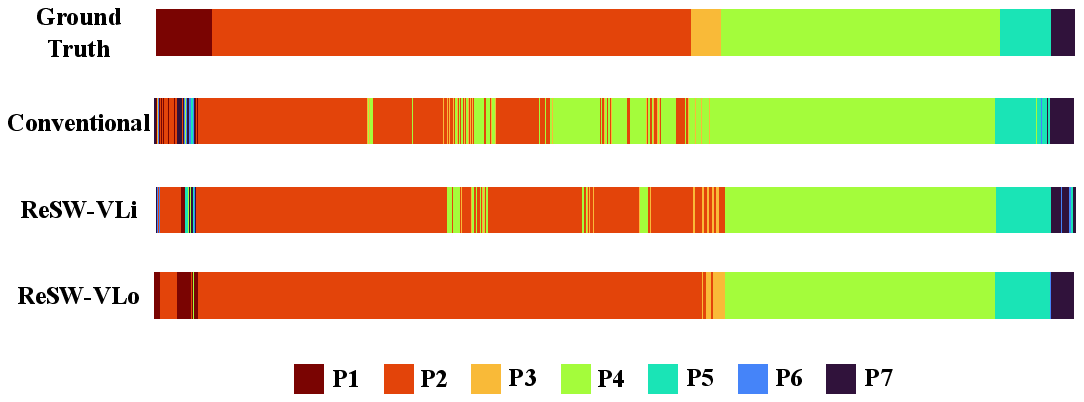}
\caption{Qualitative results of the predictions for video 50 in Cholec80 dataset.} \label{fig3}

\vspace{2\baselineskip}

\centering
\includegraphics[scale=0.65]{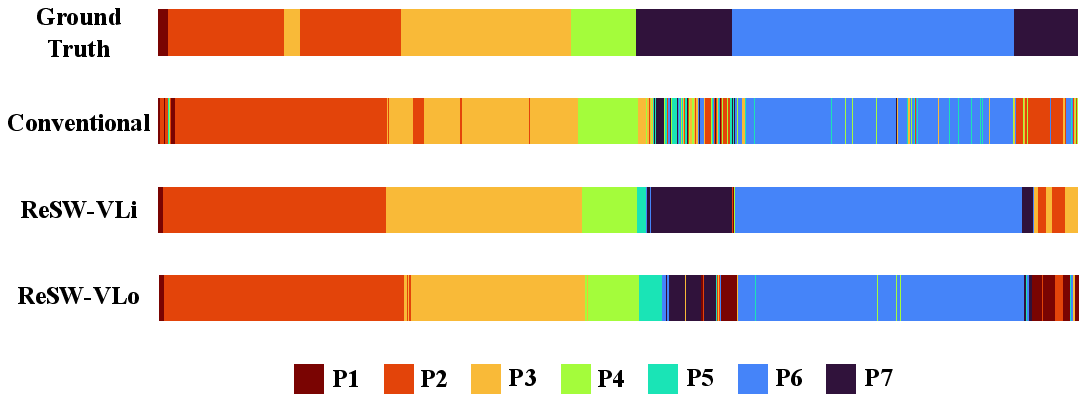}
\caption{Qualitative results of the predictions for video 15 in Autolaparo dataset.} \label{fig4}
\end{figure}

As illustrated in Figures~\ref{fig3} and \ref{fig4}, the prediction results for video50 in the test data of the Cholec80 dataset and video15 in the test data of the Autolaparo dataset are displayed, respectively. The conventional method and the proposed methods in addition to the ground truth are represented. It is evident from Figures~\ref{fig3} and \ref{fig4} that the conventional method is unstable and switches the surgical phases multiple times. In contrast, the proposed methods provide stable predictions. And we can confirm ReSW-VLo performed better on the Cholec80 dataset, while ReSW-VLi performed better on the Autolaparo dataset. 

The proposed method in this experiment utilizes CLIP's ResNet-50 for spatial feature extraction and causal TCN for time series modeling. However, it should be noted that there is potential for further enhancement of performance through the integration of alternative methods. For instance, as demonstrated in \cite{zou2023arst}, the incorporation of the Transformer decoder within the ResNet+TCN framework has been shown to enhance the accuracy by approximately 1.3 percentage points and the Jaccard Index by around 3.5 percentage points in comparison with the ResNet+TCN framework. Given that the proposed method represents an enhancement to the representation learning of spatial features, it can be extended in numerous ways.

\section{Conclusions}

In this paper, we proposed a novel approach for acquiring representations of spatial characteristics of images for the purpose of surgical phase identification. This method entailed fine-tuning the CLIP image encoder with prompt learning, a technique particularly well-suited for the recognition of surgical phases from video footage. The efficacy of this approach was substantiated by experimental findings derived from three distinct surgical phase recognition datasets. These results demonstrated that the proposed method consistently surpassed conventional methods in all three datasets.

Subsequent studies will entail the validation of the efficacy of the proposed methodology when ViT-B/16 of CLIP is employed as the image encoder and/or when Transformer is utilized for time series modeling.

%
%
%
\bibliographystyle{splncs04}
\bibliography{miccai2025}

\begin{thebibliography}{10}
\providecommand{\url}[1]{\texttt{#1}}
\providecommand{\urlprefix}{URL }
\providecommand{\doi}[1]{https://doi.org/#1}

\bibitem{czempiel2020tecno}
Czempiel, T., Paschali, M., Keicher, M., Simson, W., Feussner, H., Kim, S.T.,
  Navab, N.: Tecno: Surgical phase recognition with multi-stage temporal
  convolutional networks. In: Medical Image Computing and Computer Assisted
  Intervention--MICCAI 2020: 23rd International Conference, Lima, Peru, October
  4--8, 2020, Proceedings, Part III 23. pp. 343--352. Springer (2020)

\bibitem{czempiel2021opera}
Czempiel, T., Paschali, M., Ostler, D., Kim, S.T., Busam, B., Navab, N.: Opera:
  Attention-regularized transformers for surgical phase recognition. In:
  Medical Image Computing and Computer Assisted Intervention--MICCAI 2021: 24th
  International Conference, Strasbourg, France, September 27--October 1, 2021,
  Proceedings, Part IV 24. pp. 604--614. Springer (2021)

\bibitem{demir2023deep}
Demir, K.C., Schieber, H., Weise, T., Roth, D., May, M., Maier, A., Yang, S.H.:
  Deep learning in surgical workflow analysis: a review of phase and step
  recognition. IEEE Journal of Biomedical and Health Informatics
  \textbf{27}(11),  5405--5417 (2023)

\bibitem{deng2009imagenet}
Deng, J., Dong, W., Socher, R., Li, L.J., Li, K., Fei-Fei, L.: Imagenet: A
  large-scale hierarchical image database. In: 2009 IEEE conference on computer
  vision and pattern recognition. pp. 248--255. Ieee (2009)

\bibitem{dosovitskiy2020image}
Dosovitskiy, A., Beyer, L., Kolesnikov, A., Weissenborn, D., Zhai, X.,
  Unterthiner, T., Dehghani, M., Minderer, M., Heigold, G., Gelly, S., et~al.:
  An image is worth 16x16 words: Transformers for image recognition at scale.
  arXiv preprint arXiv:2010.11929  (2020)

\bibitem{eigen2015predicting}
Eigen, D., Fergus, R.: Predicting depth, surface normals and semantic labels
  with a common multi-scale convolutional architecture. In: Proceedings of the
  IEEE international conference on computer vision. pp. 2650--2658 (2015)

\bibitem{funke2023metrics}
Funke, I., Rivoir, D., Speidel, S.: Metrics matter in surgical phase
  recognition. arXiv preprint arXiv:2305.13961  (2023)

\bibitem{gao2021trans}
Gao, X., Jin, Y., Long, Y., Dou, Q., Heng, P.A.: Trans-svnet: Accurate phase
  recognition from surgical videos via hybrid embedding aggregation
  transformer. In: Medical Image Computing and Computer Assisted
  Intervention--MICCAI 2021: 24th International Conference, Strasbourg, France,
  September 27--October 1, 2021, Proceedings, Part IV 24. pp. 593--603.
  Springer (2021)

\bibitem{he2016deep}
He, K., Zhang, X., Ren, S., Sun, J.: Deep residual learning for image
  recognition. In: Proceedings of the IEEE conference on computer vision and
  pattern recognition. pp. 770--778 (2016)

\bibitem{hochreiter1997long}
Hochreiter, S., Schmidhuber, J.: Long short-term memory. Neural computation
  \textbf{9}(8),  1735--1780 (1997)

\bibitem{jin2017sv}
Jin, Y., Dou, Q., Chen, H., Yu, L., Qin, J., Fu, C.W., Heng, P.A.: Sv-rcnet:
  workflow recognition from surgical videos using recurrent convolutional
  network. IEEE transactions on medical imaging  \textbf{37}(5),  1114--1126
  (2017)

\bibitem{lea2016temporal}
Lea, C., Vidal, R., Reiter, A., Hager, G.D.: Temporal convolutional networks: A
  unified approach to action segmentation. In: Computer vision--ECCV 2016
  workshops: Amsterdam, the Netherlands, October 8-10 and 15-16, 2016,
  proceedings, part III 14. pp. 47--54. Springer (2016)

\bibitem{li2022ordinalclip}
Li, W., Huang, X., Zhu, Z., Tang, Y., Li, X., Zhou, J., Lu, J.: Ordinalclip:
  Learning rank prompts for language-guided ordinal regression. Advances in
  Neural Information Processing Systems  \textbf{35},  35313--35325 (2022)

\bibitem{li2024deep}
Li, Y., Zhao, Z., Li, R., Li, F.: Deep learning for surgical workflow analysis:
  a survey of progresses, limitations, and trends. Artificial Intelligence
  Review  \textbf{57}(11), ~291 (2024)

\bibitem{liu2025lovit}
Liu, Y., Boels, M., Garcia-Peraza-Herrera, L.C., Vercauteren, T., Dasgupta, P.,
  Granados, A., Ourselin, S.: Lovit: Long video transformer for surgical phase
  recognition. Medical Image Analysis  \textbf{99},  103366 (2025)

\bibitem{liu2023skit}
Liu, Y., Huo, J., Peng, J., Sparks, R., Dasgupta, P., Granados, A., Ourselin,
  S.: Skit: a fast key information video transformer for online surgical phase
  recognition. In: Proceedings of the IEEE/CVF International Conference on
  Computer Vision. pp. 21074--21084 (2023)

\bibitem{loshchilov2017decoupled}
Loshchilov, I., Hutter, F.: Decoupled weight decay regularization. arXiv
  preprint arXiv:1711.05101  (2017)

\bibitem{mikolov2010recurrent}
Mikolov, T., Karafi{\'a}t, M., Burget, L., Cernock{\`y}, J., Khudanpur, S.:
  Recurrent neural network based language model. In: Interspeech. vol.~2, pp.
  1045--1048. Makuhari (2010)

\bibitem{radford2021learning}
Radford, A., Kim, J.W., Hallacy, C., Ramesh, A., Goh, G., Agarwal, S., Sastry,
  G., Askell, A., Mishkin, P., Clark, J., et~al.: Learning transferable visual
  models from natural language supervision. In: International conference on
  machine learning. pp. 8748--8763. PmLR (2021)

\bibitem{rivoir2024pitfalls}
Rivoir, D., Funke, I., Speidel, S.: On the pitfalls of batch normalization for
  end-to-end video learning: A study on surgical workflow analysis. Medical
  Image Analysis  \textbf{94},  103126 (2024)

\bibitem{stauder2016tum}
Stauder, R., Ostler, D., Kranzfelder, M., Koller, S., Feu{\ss}ner, H., Navab,
  N.: The tum lapchole dataset for the m2cai 2016 workflow challenge. arXiv
  preprint arXiv:1610.09278  (2016)

\bibitem{twinanda2016endonet}
Twinanda, A.P., Shehata, S., Mutter, D., Marescaux, J., De~Mathelin, M., Padoy,
  N.: Endonet: a deep architecture for recognition tasks on laparoscopic
  videos. IEEE transactions on medical imaging  \textbf{36}(1),  86--97 (2016)

\bibitem{vaswani2017attention}
Vaswani, A., Shazeer, N., Parmar, N., Uszkoreit, J., Jones, L., Gomez, A.N.,
  Kaiser, {\L}., Polosukhin, I.: Attention is all you need. Advances in neural
  information processing systems  \textbf{30} (2017)

\bibitem{wang2022autolaparo}
Wang, Z., Lu, B., Long, Y., Zhong, F., Cheung, T.H., Dou, Q., Liu, Y.:
  Autolaparo: A new dataset of integrated multi-tasks for image-guided surgical
  automation in laparoscopic hysterectomy. In: International Conference on
  Medical Image Computing and Computer-Assisted Intervention. pp. 486--496.
  Springer (2022)

\bibitem{zou2023arst}
Zou, X., Liu, W., Wang, J., Tao, R., Zheng, G.: Arst: auto-regressive surgical
  transformer for phase recognition from laparoscopic videos. Computer Methods
  in Biomechanics and Biomedical Engineering: Imaging \& Visualization
  \textbf{11}(4),  1012--1018 (2023)

\end{thebibliography}

\end{document}